\documentclass[10pt,twocolumn,letterpaper]{article}

\usepackage{cvpr}
\usepackage{times}
\usepackage{epsfig}
\usepackage{graphicx}
\usepackage{amsmath}
\usepackage{amssymb}

\usepackage[pagebackref=true,breaklinks=true,colorlinks,bookmarks=false]{hyperref}

\cvprfinalcopy % *** Uncomment this line for the final submission

\def\cvprPaperID{****} % *** Enter the CVPR Paper ID here

\ifcvprfinal\fi
\begin{document}

%%%%%%%%% TITLE
\title{A Review of Vision-Language Models and their Performance on the Hateful Memes Challenge}

\author{Bryan Zhao$^*$, Andrew Zhang$^*$, Blake Watson$^*$, Gillian Kearney$^*$, Isaac Dale$^*$\\
Georgia Institute of Technology\\
{\tt\small bzhao90, azhang377, bwatson60, gkearney3, idale7 @gatech.edu}
}

\maketitle
%\thispagestyle{empty}

%%%%%%%%% ABSTRACT
\begin{abstract}
Moderation of social media content is currently a highly manual task, yet there is too much content posted daily to do so effectively. With the advent of a number of multimodal models, there is the potential to reduce the amount of manual labor for this task. In this work, we aim to explore different models and determine what is most effective for the Hateful Memes Challenge \cite{HM-Challenge}, a challenge by Meta designed to further machine learning research in content moderation. Specifically, we explore the differences between early fusion and late fusion models in classifying multimodal memes containing text and images. We first implement a baseline using unimodal models for text and images separately using BERT and ResNet-152, respectively. The outputs from these unimodal models were then concatenated together to create a late fusion model. In terms of early fusion models, we implement ConcatBERT, VisualBERT, ViLT, CLIP, and BridgeTower. It was found that late fusion performed significantly worse than early fusion models, with the best performing model being CLIP which achieved an AUROC of 70.06. The code for this work is available at \hyperlink{https://github.com/bzhao18/CS-7643-Project}{https://github.com/bzhao18/CS-7643-Project}.
\end{abstract}

%%%%%%%%% BODY TEXT
\section{Introduction \& Motivation}

\begin{figure}[t]
\begin{center}
   \includegraphics[width=1\linewidth]{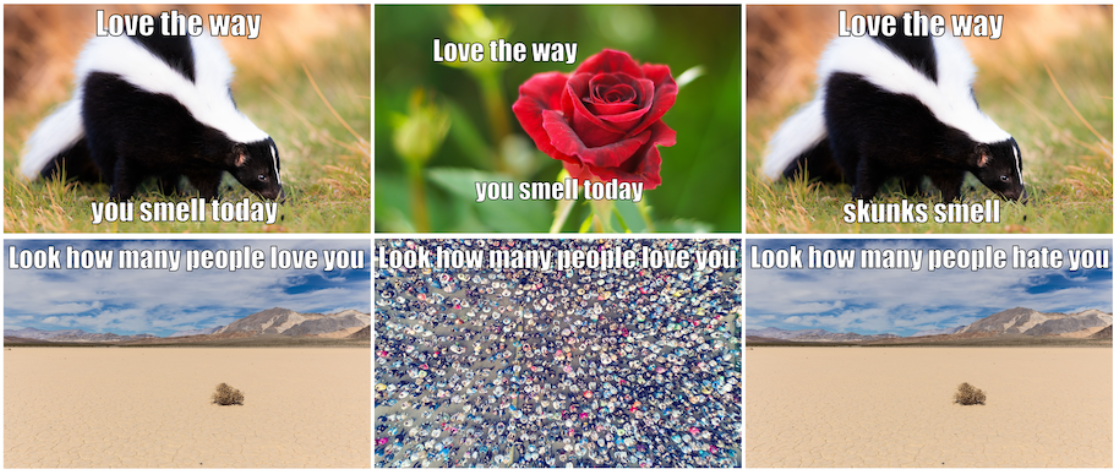}
\end{center}
   \caption{A couple of examples memes from the Hateful Memes Challenge \cite{HM-Challenge}. Notice the memes in the left column are hateful, the memes in the middle column are not hateful, and the memes in the right column have the same images as the left column, but due to a change in the text, the memes are not hateful.}
\label{fig:memes}
\end{figure}

Today the use of the Internet has become a part of everyday life, and the world is more interconnected than ever. However, in a world where people can post and send messages at the push of a button, hateful content and speech often go uncensored. There is too much data created every second to be moderated manually, yet large social media companies must filter content that is harmful to its users. The difficulty lies in determining what content is hateful and what content is completely acceptable. This is compounded by the fact that communication comes in various different modes and in various different contexts which fundamentally change the meaning that is conveyed. Based on this, memes have been identified as a prime target for developing models that can automate content moderation as the joking nature of memes along with the complex interaction between a meme's image and text make it particularly difficult to classify as hateful or acceptable as demonstrated in Figure \ref{fig:memes}.

Current hate speech classification is done in a hybrid model. There are a number of ways that automate the detection of hate speech. Pre-flagged hate speech is uniquely hashed so when uploads take place, they can be taken down without human verification \cite{ContentModeration}. Other methods include user flagging and content moderator review where users and content moderators review and determine if content is hateful. These content moderators are constantly inundated with horrible content which causes many to be driven to psychological harm including extreme drug use and suicidal thoughts \cite{TraumaFloor}. This heavy reliance on human content moderators is an extreme ethical issue. 
		
Creating a model that can correctly classify hate speech protects users and moderators alike from the harmful effects. A successful model will enable users have a good online experience while protecting content moderators and the victims of the hate speech that is being moderated. 

This project will utilize the dataset from the Hateful Memes Challenge, a dataset provided by Meta that was created to further research in content moderation \cite{HM-Challenge}. The dataset is comprised of a set of 10,000 memes each with their image, text, and label (either hateful or not) split into development, test, and training sets. Labels were annotated based off the following definition of hate: 
\begin{quote}
    ``A direct or indirect attack on people based on characteristics, including ethnicity, race, nationality, immigration status, religion, caste, sex, gender identity, sexual orientation, and disability or disease. We define attack as violent or dehumanizing (comparing people to non-human things, e.g. animals) speech, statements of inferiority, and calls for exclusion or segregation. Mocking hate crime is also considered hate speech.''
\end{quote}

The created memes are based off of real memes from the Internet, but images were ultimately sourced from Getty Images to prevent licensing issues. Any memes that contained slurs were immediately removed from the collected dataset as they are clearly identifiable as hateful. To further encourage a multimodal approach, benign confounders were added into the set. Keeping the same set of text but changing the image can change the meaning of the meme. For example, “you smell great” with a rose in the background versus a skunk has remarkably different meanings despite using the same text. The total composition of the dataset is as follows:

\begin{quote}
    ``The dev and test set are fully balanced, and are comprised of memes using the following percentages: 40\% multimodal hate, 10\% unimodal hate, 20\% benign text confounder, 20\% benign image confounder, 10\% random non-hateful \cite{HM-Challenge}.''
\end{quote}

\section{Related Works}
\begin{figure}[t]
\begin{center}
   \includegraphics[width=1\linewidth]{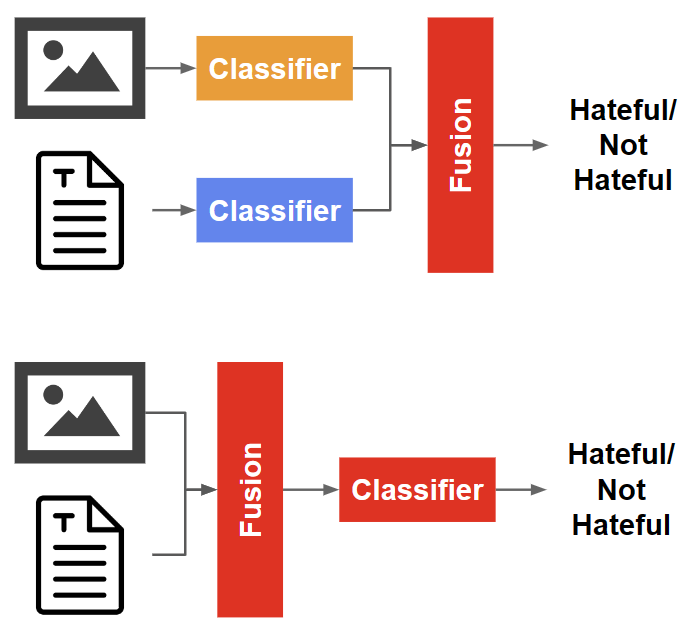}
\end{center}
   \caption{Comparison of late fusion (top) versus early fusion (bottom) models for image and text data.}
\label{fig:fusion}
\end{figure}

Based on the original competition paper for the Hateful Memes Challenge \cite{HM-Challenge} and other investigations into multimodal models \cite{EarlyLateFusion}, early fusion models, or models that attempt to combine modalities early on before classification, are much more effective than late fusion models, or models that attempt to classify the content for each modality before fusing results. The difference between the two can be seen in Figure \ref{fig:fusion}. In this work, we attempt to confirm their findings.

The general methods of the top submissions to the Hateful Memes Challenge \cite{Zhu-1st}, \cite{Muennighoff-2nd}, \cite{Velioglu-3rd}, \cite{Lippe-4th} were to first use varying forms of data augmentation and feature extraction to generate more information on each of the images and text. Next, these were then used to fine-tune various pretrained vision-language models for the task of hateful meme classification, and these models included VL-BERT \cite{VLBERT}, VisualBERT \cite{VisualBERT}, ViLBERT \cite{ViLBERT}, UNITER \cite{UNITER}, and OSCAR \cite{OSCAR}. Lastly, ensemble learning combined predictions from all of the models that were implemented. A similar approach in this work will be followed using current SOTA models. 

%-------------------------------------------------------------------------

\section{Approach}
We experimented with multiple types of models and various methods of multimodal fusion to obtain classifications on memes. Our implementations include three baseline models from the original challenge \cite{HM-Challenge}: Basic Late Fusion, ConcatBERT, and VisualBERT, which was also used in a winning solution \cite{Velioglu-3rd} to the 2020 Hateful Memes Challenge. Note that we did not use the implementations provided by Meta’s MMF framework \cite{MMF}, and instead opted to use the Transformers API from HuggingFace \cite{HuggingFace}. We also explored additional multimodal models with different fusion methods using CLIP \cite{CLIP} and BridgeTower \cite{BridgeTower}.

\begin{figure*}[t]
\begin{center}
    \includegraphics[width=0.8\linewidth]{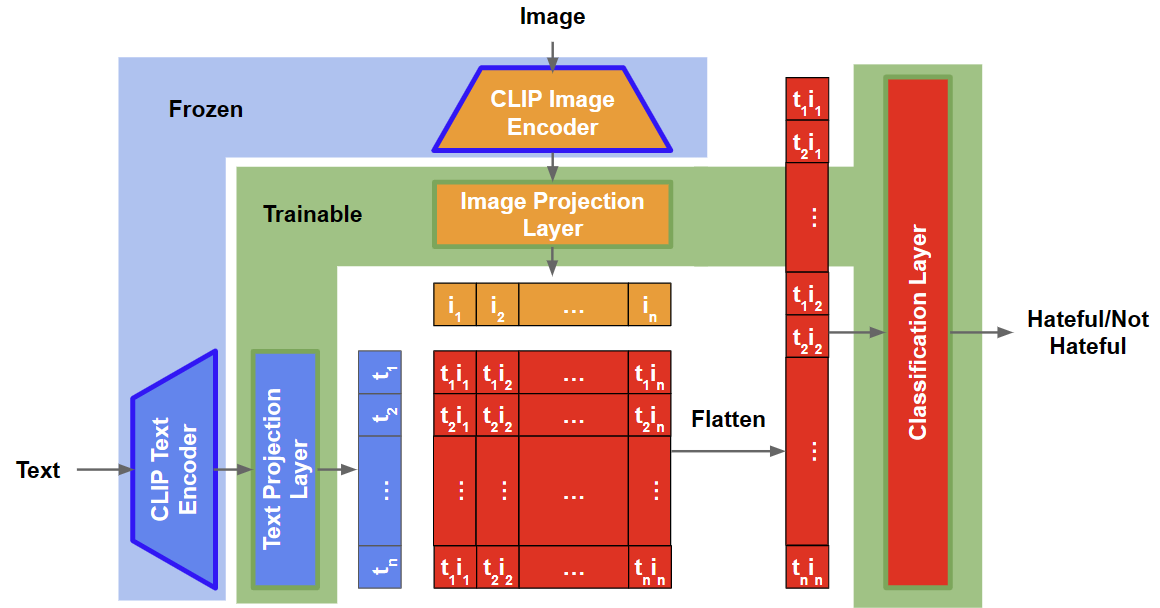}
\end{center}
   \caption{Architecture using CLIP to classify hateful memes based on Hate-CLIPper \cite{hateclipper}. Two versions were implemented: one with the trainable text projection and image projection trainable layers and one without.}
\label{fig:clip}
\end{figure*}

\subsection{Data Preparation and Manipulation}
The Hateful Memes dataset was split into multiple parts labeled “train”, “dev\_seen”, “dev\_unseen”, “test\_seen”, and “test\_unseen”. The “test\_seen” and “test\_unseen” datasets were used for evaluation purposes in the competition and did not include labels. As such, this data was omitted from our experiment. The “train” dataset consisting of 8500 labeled memes was used to train and fine-tune the implementations while the “dev\_seen” and “dev\_unseen” datasets were combined to obtain a total of 1040 memes for evaluation.
All images were accompanied by the extracted meme text and labeled as hateful/non-hateful. The images were resized to a standard 300x300 and center-cropped to 256x256 in order to omit most of the top/bottom text present in the meme images and focus on the image content.

\subsection{Baseline Unimodal and Late Fusion Models}
As a baseline, two unimodal models were implemented using a pretrained BERT model (Bidirectional Encoder Representations from Transformers) \cite{BERT} from HuggingFace to classify the text of the memes and a pretrained ResNet-152 Model \cite{ResNet} to classify the images. Both models were further fine-tuned on the Hateful Memes data. Furthermore, a basic late fusion model was created by combining the normalized logits from both unimodal models by summing to form a singular set of logits for classification. 

\subsection{ConcatBERT}
We also implemented a basic early fusion model from the original challenge \cite{HM-Challenge} consisting of a pre-trained BERT from HuggingFace \cite{BERT} to generate text features and ResNet-152 pre-trained on ImageNet from PyTorch \cite{ResNet} to generate image features. The BERT model from HuggingFace was pre-trained for a multitude of NLP tasks, however, only the base BERT model was used to output the raw hidden states and obtain the textual features. The 768-dimensional text feature vector output by BERT and the 2048-dimensional image feature vector from ResNet-152 are concatenated to create a 2816-dimensional multimodal feature vector which is then passed through a 2-layer feed-forward network that serves as the classification head. The training process mainly focused on the feed-forward layers, while the weights for both the BERT and ResNet models were frozen. The FFN consists of a BatchNorm layer followed by two fully connected layers with ReLU nonlinearity and Dropout in between. It was observed that the model initially had overfitting issues and the inclusion of the BatchNorm and Dropout layers along with explicit L2 weight penalties improved generalization. 

\begin{figure*}[t]
\begin{center}
    \includegraphics[width=0.8\linewidth]{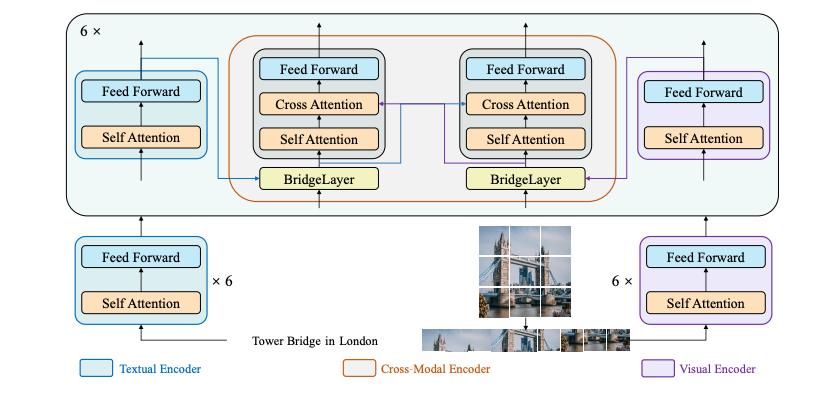}
    \caption{Architecture of the base BridgeTower model with multiple layers of unimodal encoders interacting with cross-modal encoders. The pooled output from the final layer of cross-modal encoders was used to obtain the a multimodal feature vector.  }
\end{center}
\label{fig:short}
\end{figure*}

\subsection{CLIP}
CLIP \cite{CLIP} is a state of the art vision-language model that was originally developed by OpenAI. It's original pretraining task was to match a caption with an image. It achieves this by jointly learning image and text encodings and projecting them into a shared latent space. CLIP was specifically selected based on an implementation called Hate-CLIPper which achieved SOTA performance with an AUROC of 0.858 \cite{hateclipper}.

Two main versions using CLIP were developed as shown in Figure \ref{fig:clip}. The first model used CLIP with only a classification head. The CLIP image and text encoders were frozen and used to produce embedding vectors. The outer product is taken to obtain a multimodal fusion matrix. This matrix is flattened and input into a shallow feed-forward network. The second approach uses a similar approach to Hate-CLIPper where shallow fully-connected networks are inserted after the text and image encoders \cite{hateclipper}. After the classification head was trained, the encoders were unfrozen, and the entire model was fine-tuned with a low learning rate.

A key issue that was the model overfitting on the testing data. This was a common issue across all models due to the fact that the dataset is small relative to the number of parameters in these pretrained models. In order to reduce overfitting as much as possible, standard practices such as dropout, weight decay, data shuffling, and early stopping were employed. Furthermore, random image transforms were used including random horizontal and vertical flipping, random cropping, and color jitter.

Furthermore, another issue was computational limitations as the largest version of CLIP could not be fine-tuned on a single GPU on Google Colab Pro. Therefore, a smaller version of CLIP called ViT-Base-Patch32 was used rather than the larger version ViT-Large-Patch14 which was used in Hate-CLIPper.

\subsection{BridgeTower}

BridgeTower \cite{BridgeTower} builds on the Two-Tower architecture and combines multiple layers of textual and visual unimodal encoders and interweaves them with multiple layers of cross-modal encoders, instead of simply feeding the last layer outputs of unimodal encoders into a singular cross-modal encoder. BridgeTower allows for a more comprehensive fusion process, where the image and textual features are able to interact at each layer. Since the problem of identifying hateful memes is much more nuanced than just image-to-text matching, the detailed fusion process could potentially better capture the complicated relationships between the meme text and image content. 

While the BridgeTower implementation from huggingface \cite{BridgeTower} was pre-trained for masked language modeling and image-to-text matching, our implementation uses the base BridgeTower model without the classification heads for specific downstream tasks. The pooled outputs of the final layer of cross-modal encoders is used to obtain a multimodal feature vector. This feature vector is then passed into a feed-forward network instead, consisting of two fully-connected layers with ReLU nonlinearity and Dropout. The pre-trained BridgeTower model is frozen and the training process is focused on the FFN classification head. The initial model had overfitting issues and the addition of explicit L2 weight penalties with Dropout somewhat improved generalization.

\subsection{VisualBERT}
A model using VisualBERT \cite{VisualBERT} was also implemented based on the original Hateful Memes paper \cite{HM-Challenge}. This model uses visual embeddings in the form of bounding boxes and features from an Faster R-CNN-based object detector. In the original implementation, these features were generated using a COCO pre-trained Faster R-CNN model from the depreciated Detectron codebase and was modified to the Detectron2 version \cite{detectron2}. The text embeddings are obtained using BERT, and both embeddings are then passed into VisualBERT to obtain a multimodal embedding. This model, in theory, has improved recognition of positional text in images, while maintaining the text recognition of the baseline BERT model. The multimodal embedding is then passed through a classification head consisting of a feed-forward network with ReLU nonlineaities to obtain the final classifications. 

While VisualBERT originally uses a pre-trained Detectron model on the COCO dataset, it tends to overproduce features which are smaller than needed, with most images having roughly 50 bounding boxes. We did attempt to create a custom Detectron2 model that would produce fewer visual features, but the balance between the size of bounding boxes and number of features extracted proved difficult to develop. Due to the number of visual embeddings, the size of the input to the VisualBERT model became unexpectedly large. This severely impacted the runtime, with a single epoch taking over 30 minutes on a high end graphics card. This issue was further exacerbated by the Fast R-CNN model, which was completing slower than expected due to the increased number of features being output. This made it difficult to train and modify hyperparameters.

\subsection{ViLT}
ViLT \cite{ViLT} simplifies the vision-and-language model using the transformer to extract visual features instead of a deep visual embedding. Specifically the visual feature layer of ViLT encodes them in a way that is as light as encoding the text, and does not involve convolution. Instead of using an object detection model like Faster R-CNN for its visual feature extractor, ViLT uses a patch projection, taking 32x32 flattened patches and using them for image embeddings. ViLT focuses on the modality interaction layers with the minimal amount of encoding before fusing together. This means that the multimodal features are likely to be well trained on the interaction between text and image input.

The ViLT implementation from HuggingFace was pretrained on Image Text Matching, Masked Language Modeling and Whole Word Masking. Our implementation uses the base ViLT model without the heads for downstream tasks to obtain a multimodal feature vector to pass into a feed-forward network, consisting of two linear layers with ReLU nonlinearity and Dropout. The pre-trained ViLT model’s weights are frozen and the training process is focused on the binary classification head. While ideally the multimodal features encoder weights could have been trained as well, unfortunately we ran into memory issues with CUDA when attempting to train this. The initial model had overfitting issues, which was expected, given the complexity of the ViLT model and the number of multimodal features. The addition of explicit L2 weight penalties with Dropout somewhat improved generalization.

\section{Experiments and Results}

\begin{table*}
\begin{center}
\begin{tabular}{|l|c|c|c|c|}
\hline
Model & Training Accuracy & Training AUROC & Validation Accuracy & Validation AUROC \\
\hline\hline
Baseline Late Fusion & 0.979 & 0.9736 & 0.549 & 0.5453 \\
\hline
 Unimodal - BERT & 0.8820 & 0.8761 & 0.5586 & 0.5586 \\
\hline
 Unimodal - ResNet & 0.9369 & 0.9312 & 0.5124 & 0.5098 \\
\hline
ConcatBERT & 0.8096 & 0.7707 & 0.5820 & 0.5791 \\
\hline
CLIP & 0.8896 & 0.8724 & \textbf{0.6492} & \textbf{0.7006} \\
\hline
Hate-CLIPper & 0.9438 & 0.9392 & 0.6384 & 0.6432 \\
\hline
BridgeTower & 0.7171 & 0.6523 & 0.6200 & 0.6192 \\
\hline
VisualBERT & 0.6834 & 0.6912 & 0.5723 & 0.5802 \\
\hline
ViLT & 0.7273 & 0.6695 & 0.5960 & 0.5932 \\
\hline
\end{tabular}
\end{center}
\caption{Training and validation results for each model}
\label{tab:results}
\end{table*}

\subsection{Loss and Metrics}
Across all models, the area under the receiver operating characteristic (AUROC) was used as the evaluation metric, as in the original Hateful Memes Challenge, and is recorded alongside the accuracy. Additionally, all models were trained using Cross Entropy Loss with the Adam optimizer.

\subsection{Overall Results}
The training and validation accuracy and AUROC for each of the models are listed in Table \ref{tab:results}. Overall, the best performance was achieved using the base CLIP model with a classification head which obtained a validation AUROC of 0.7006. This performance is indeed comparable to the initial results of the other models found in \cite{HM-Challenge}. Overall, it seems that the early fusion models including ConcatBERT, CLIP, BridgeTower, VisualBERT, and ViLT are able to achieve significantly better results than the basic unimodal and late fusion models. These results are consistent with other findings on the effectiveness of early and late fusion methods for multimodal classification \cite{EarlyLateFusion}. 

\subsection{Unimodal and Late Fusion Models}
The naive unimodal BERT and ResNet models along with the baseline late fusion models achieve moderate success in distinguishing hateful memes, however they perform much worse than the more complex models. Notably, it seems that on its own, the textual information is much more significant than the image content in determining hatefulness. The unimodal text-based BERT classifier performs significantly better than the unimodal ResNet visual classifier with an AUROC of 0.5586 compared to 0.5098. However, when the two models are combined using basic late fusion, the performance actually worsens when compared to the unimodal textual classifier, achieving only an AUROC of 0.5453. This suggests that the naive method of late fusion is unable to properly capture the interactions between the text and image content and evaluate the overall intent of the meme altogether. Additionally, we found that all three models had severe overfitting issues and generalized poorly.

\subsection{Early Fusion Models}

The early fusion models including ConcatBERT, CLIP, BridgeTower, VisualBERT and ViLT all achieve significantly better results than the naive unimodal and late fusion models.

The baseline ConcatBERT achieves an AUROC of 0.5791, slightly outperforming the baseline late fusion model. However, it falls short of the implementation found in \cite{HM-Challenge}. This difference is likely due to the fact that only the classification head consisting of a feed-forward network was trained on the dataset, while the feature extractors consisting of BERT and ResNet152 were frozen to keep the training process simple and minimize computation costs. Additionally, we used slightly different implementations of BERT and ResNet-152 which may have been fine-tuned differently. However, the comparisons between our implementations are consistent with the findings from \cite{HM-Challenge}, where the basic early fusion ConcatBERT achieves a slight improvement over late fusion models.

The VisualBERT model achieves slightly better results than ConcatBERT with an AUROC of 0.5802 using a more sophisticated method of fusion. However, the results fall short of both the baseline implementaton in \cite{HM-Challenge} and the winning solution \cite{Velioglu-3rd}. This is likely due to a multitude of trade-offs to account for the limited computation power available. Due to the excessive time required to train the model, very little fine-tuning was performed, and only the final classification head was trained while all other weights for the encoders and VisualBERT were frozen. Further fine-tuning on the encoders and VisualBERT itself could likely yield better results. Despite this, the more complex fusion provided by VisualBERT improves over the basic fusion method of concatenation in ConcatBERT.

The baseline CLIP model achieved an impressive AUROC of 0.7006, while our implementation of Hate-CLIPper using additional fully-connected layers alongside the text and image encoders only achieved an AUROC of 0.6432. We found that the addition of the fully-connected layers in Hate-CLIPper resulted in severe overfitting, as the implementation saw extremely high training accuracies and AUROC, despite the addition of Dropout layers, random image transforms, and explicit L2 weight decay. The added layers seem to create a much more complex model that leads to the observed generalization issues. Additionally, similar to VisualBERT, our results fall short of the implementation in the original paper \cite{hateclipper}. This is likely due to the fact that we used a smaller, more lightweight version of CLIP, ViT-Base-Patch32, in order to account for limited computation power and perform more fine-tuning. The faster training times afforded by the lightweight model allowed for additional fine-tuning alongside the use of a multimodal fusion matrix, which could explain why CLIP still saw significant improvement over the other tested models. The

Both BridgeTower and ViLT were able to achieve respectable AUROCs of 0.6192 and 0.5932 respectively, again performing significantly better than the late fusion and unimodal models. This suggests that the increased interaction allowed by the multimodal and unimodal encoders in BridgeTower and the ViLT encoding with a larger modality interaction seem to better capture the relationship between the meme's text and image content. However, both implementations fall short of CLIP. This is likely due to the fact that BridgeTower and ViLT are larger models, requiring more time and computing power to train and run. This left less time for hyperparamter tuning and further fine-tuning which could have improved results. 

Overall, our findings are consistent with \cite{HM-Challenge} and \cite{EarlyLateFusion}, showing that early fusion models significantly outperform late fusion models when dealing with multimodal input. Additionally, we found that the methods of fusion also have an impact on the effectiveness of the model, where the more complex fusion methods used in CLIP, BridgeTower, and ViLT that allow for more detailed interaction between the image and textual features outperform the basic fusion method of concatenation used in ConcatBERT.

\section{Experience}

\subsection{Challenges}
We faced a number of challenges over the course of this project, which mainly consisted of computing limitations, dataloader issues, and our inability to fine-tune feature extractors. Due to the nature of a number of the models that we used, both our local runtime environments as well as Colab environments ran into computing power limitations in which we were unable to train certain model parameters or increase the batch size. We did attempt to mitigate runtime issues by preloading the image data into HDF5 binary files. This allowed for slightly faster load times, but the batch size remained relatively small due to the same computing limitations. The last major issue that we encountered was the issue of being unable to fine-tune feature extractors. Nearly all the models that were utilized a multimodal feature extractor, which was pretuned on other datasets, often MS COCO or ImageNet. Ideally, we would fine-tune this with the Hateful Memes dataset, but we ran out of CUDA memory in these scenarios. 

\subsection{Changes in Approach}
There are many changes that could be made in the future to improve upon our current approach. Due to the computation issues previously mentioned, we made a few shortcuts and simplifications to the models that were used to attempt to minimize training time and overall runtimes. We were forced to freeze model parameters on most of the larger models, as attempting to load the models parameters into the optimizer resulted in cuda memory allocation errors. This meant that we were only training the classification head parameters, which was effective to some degree but less ideal than training the entire model. We were also unable to fine-tune a number of the larger models, mainly due to time constraints caused by the same computing limitations. Many of the models could be improved by further fine-tuning the pre-trained encoders and fusion models on the hateful memes themselves. 

We would have also liked to further investigate different methods of pre-processing that images and/or augmenting the dataset. Many of the winning solutions in the original competition achieved their results due to additional pre-processing of images and introducing additional data for fine-tuning.

\subsection{Project Success}
Overall our project successfully implemented different multimodal models to classify hateful memes based on both its text and image content. We saw some shortcomings in a few of our models compared to the original implementations and competition submissions, most of which can be attributed to our lack of computing resources leading to less fine-tuning. Despite these shortcomings, we were able to make comparisons between the different approaches. We were able to demonstrate that many early fusion models can achieve much better results when compared to late fusion models.

%-------------------------------------------------------------------------

{\small
\bibliographystyle{ieee_fullname}
\bibliography{egbib}
}

\end{document}